\documentclass[11pt,a4paper]{article}

\newcommand{\stddev}[2]{\ensuremath{#1_{\color{darkgray}{\pm #2}}}} 
\usepackage{cancel}

\usepackage{latexsym}
\usepackage{tikz-dependency}
\usepackage{graphicx}
\usepackage{amsmath}  % AMS mathematical facilities for LaTeX.
\usepackage{amsthm}   % Typesetting theorems (AMS style).
\usepackage{amssymb} % changed from amsfonts because it includes more

\usepackage{fixltx2e}
\usepackage{booktabs}
\usepackage{tablefootnote}
\usepackage{makecell}
\usepackage{multirow}
\usepackage[export]{adjustbox}
\usepackage{dsfont}
\usepackage{fancyhdr}
\usepackage{subcaption} % subfig and caption formating
\usepackage{xcolor} % better than color
\usepackage{enumitem} % supercedes enumerate, nice lists
\usepackage{listings} %show code, better than verbatim or lstings
\usepackage{mathtools} % more symbols, eqs numbering etc, good stuff
\usepackage{csquotes} % support for good quotations
\usepackage{cases} % cases eq numbering
\usepackage{xspace}
\usepackage{comment}
\usepackage{pifont}% http://ctan.org/pkg/pifont
\usepackage{fixmath}
\usepackage{algorithm2e}
\usepackage{naaclhlt2019}
\usepackage{times}
\usepackage{latexsym}
\usepackage{cleveref} %best refs
\aclfinalcopy % Uncomment this line for the final submission
\def\aclpaperid{1081} %  Enter the acl Paper ID here

\renewcommand{\vec}[1]{\mathbf{#1}}

\title{Viable Dependency Parsing as Sequence Labeling}

\author{Michalina Strzyz \qquad David Vilares \qquad Carlos G\'omez-Rodr\'iguez\\
  Universidade da Coru\~na, CITIC \\
  FASTPARSE Lab, LyS Research Group, Departamento de Computaci\'on \\
  Campus de Elvi\~na, s/n, 15071 A Coru\~na, Spain\\
  {\tt \{michalina.strzyz,david.vilares,carlos.gomez\}@udc.es}}
  
\date{}

\begin{document}
\maketitle
\begin{abstract}
We recast dependency parsing as a sequence labeling problem, 
exploring several encodings of dependency trees as labels. While dependency parsing by means of sequence labeling had been attempted in existing work, results suggested that the technique was impractical. We show instead that with a conventional \textsc{bilstm}-based model it is possible to obtain fast and accurate parsers. These parsers are conceptually simple, not needing traditional parsing algorithms or auxiliary structures. However, experiments on the PTB and a sample of UD treebanks show that they provide
a good speed-accuracy tradeoff, with results competitive with more complex approaches.
\end{abstract}

\section{Introduction}

The application of neural architectures to syntactic parsing, and especially the ability of long short-term memories (LSTMs) to obtain context-aware feature representations \citep{hochreiter1997long}, has made it possible to
parse natural language with conceptually simpler models than before. 
For example, in dependency parsing,
the rich feature models with dozens of features used in transition-based approaches 
\citep{zhang-nivre:2011:ACL-HLT2011} can be simplified 
when using feedforward neural networks \citep{chen-manning:2014:EMNLP2014}, and even more with BiLSTM architectures \citep{TACL885}, where in fact two positional features can suffice \citep{shi-huang-lee:2017:EMNLP2017}. 
Similarly, in graph-based approaches,
\citet{dozat-iclr} have shown that an arc-factored model can achieve state-of-the-art accuracy, without the need for the higher-order features used in systems like \citep{koo-collins:2010:ACL}.

In the same way,
neural feature representations have made it possible to
relax the need for structured representations. This is the case of
sequence-to-sequence models that translate sentences into linearized 
trees, which were first applied to constituent 
\citep{vinyals2015grammar} and later to dependency parsing \citep{wiseman-rush:2016:EMNLP2016,zhang-EtAl:2017:EMNLP20173,li-EtAl:2018:C18-13}. 
Recently, \citet{GomVilEMNLP2018} have shown that sequence labeling models, where each word is associated with a label (thus simpler than sequence to sequence, where the
mapping from input to output is not one to one)
can learn constituent parsing.

\paragraph{Contribution} We show that sequence labeling is
useful for dependency parsing, in contrast to previous work \cite{Spoustova,li-EtAl:2018:C18-13}.
We explore four different encodings to represent dependency trees for a sentence of length $n$ as a set of $n$ labels associated with its words. We then use these representations to perform dependency parsing with an off-the-shelf sequence labeling model. The results show that we produce models with an excellent speed-accuracy tradeoff, without requiring any explicit parsing algorithm or auxiliary structure (e.g. stack or buffer). The source code is available at \url{https://github.com/mstrise/dep2label}

\section{Parsing as sequence labeling}

Sequence labeling is a structured prediction problem where a single output label is generated for every input token. This is the case of tasks such as PoS tagging, chunking or named-entity recognition, for which different approaches obtain accurate results \cite{brill1995transformation,ramshaw1999text,reimers2017reporting}. 

On the contrary, previous work on dependency parsing as sequence labeling is vague and reports results that are significantly lower than those provided by transition-, graph-based or sequence-to-sequence models \cite{Dyer2015,TACL885,dozat-iclr,zhang2017dependency}. \newcite{Spoustova} encoded dependency trees using a relative PoS-based scheme to represent the head of a node, to then train an averaged perceptron. They did not provide comparable results, but claimed that the accuracy was between 5-10\% below the state of the art in the pre-deep learning era. Recently, \citet{li-EtAl:2018:C18-13} used a relative positional encoding of head indexes with respect to the target token. This is used to train Bidirectional LSTM-CRF sequence-to-sequence models \cite{huang2015bidirectional}, that make use of sub-root decomposition. They compared their performance against an equivalent BiLSTM-CRF labeling model. 
The reported UAS for the sequence labeling model was 87.6\% on the Penn Treebank, more than 8 points below the current best model \cite{MaStackPointer}, concluding that sequence-to-sequence models are required to obtain competitive results. 

In this work, we show that these results can be clearly improved if simpler architectures are used.

\section{Encoding of trees and labels}

Given a sentence $w_1 \ldots w_n$, we associate the words with nodes $\{0, 1, \ldots, n\}$, where the extra node $0$ is used as a dummy root for the sentence. A dependency parser will find a set of labeled relations encoded as edges of the form $(h,d,l)$, where $h \in \{0, 1, \ldots, n\}$ is the head, $d \in \{1, \ldots, n\}$ the dependent, and $l$ a dependency label. The resulting dependency graph must be acyclic and such that each node in $\{1, \ldots, n\}$ has exactly one head, so it will be a directed tree rooted at node $0$.

\begin{figure}[t]
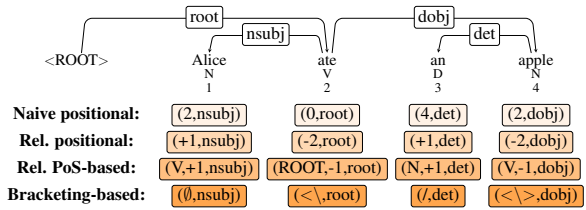

\centering
\begin{adjustbox}{max width=\columnwidth}
\begin{dependency}[label theme = default]
   \begin{deptext}[column sep=1em] 
     \larger{$<$ROOT$>$} \&  \larger{Alice} \&  \larger{ate} \&  \larger{an} \&  \larger{apple}\\
       \& N \& V \& D \& N \\
        \& 1 \& 2 \& 3 \& 4 \\ 
        \\
       \Large{\textbf{Naive positional:}} 
       \& \Large{(2,nsubj)} \& \Large{(0,root)} \& \Large{(4,det)} \& \Large{(2,dobj)} \\
       \\
       \Large{\textbf{Rel. positional:}}
       \& \Large{(+1,nsubj)} \& \Large{(-2,root)} \& \Large{(+1,det)} \& \Large{(-2,dobj)} \\
       \\
       \Large{\textbf{Rel. PoS-based:}}
       \& \Large{(V,+1,nsubj)} \& \Large{(ROOT,-1,root)} \& \Large{(N,+1,det)} \& \Large{(V,-1,dobj)} \\
       \\
       \Large{\textbf{Bracketing-based:}}
       \&  \Large{($\emptyset$,nsubj)} \& \Large{(\textless \textbackslash,root)} \& \Large{(/,det)} \& 
       \Large{(\textless \textbackslash \textgreater,dobj)} \\
   \end{deptext}
   
   \huge{\depedge{3}{2}{nsubj}}
    \huge{\depedge{3}{5}{dobj}}
    \huge{\depedge{5}{4}{det}}
     \huge{\depedge{1}{3}{root}}

    \wordgroup[group style={fill=orange!10, draw=black, inner sep=.2ex}]{5}{2}{2}{a0}
     \wordgroup[group style={fill=orange!10, draw=black, inner sep=.2ex}]{5}{3}{3}{a1}
     \wordgroup[group style={fill=orange!10, draw=black, inner sep=.2ex}]{5}{4}{4}{a2}
     \wordgroup[group style={fill=orange!10, draw=black, inner sep=.2ex}]{5}{5}{5}{a3}
     
     \wordgroup[group style={fill=orange!30, draw=black, inner sep=.2ex}]{7}{2}{2}{a4}
     \wordgroup[group style={fill=orange!30, draw=black, inner sep=.2ex}]{7}{3}{3}{a5}
     \wordgroup[group style={fill=orange!30, draw=black, inner sep=.2ex}]{7}{4}{4}{a6}
     \wordgroup[group style={fill=orange!30, draw=black, inner sep=.2ex}]{7}{5}{5}{a7}
     
     \wordgroup[group style={fill=orange!50, draw=black, inner sep=.2ex}]{9}{2}{2}{a4}
     \wordgroup[group style={fill=orange!50, draw=black, inner sep=.2ex}]{9}{3}{3}{a5}
     \wordgroup[group style={fill=orange!50, draw=black, inner sep=.2ex}]{9}{4}{4}{a6}
     \wordgroup[group style={fill=orange!50, draw=black, inner sep=.2ex}]{9}{5}{5}{a7}
     
     \wordgroup[group style={fill=orange!70, draw=black, inner sep=.2ex}]{11}{2}{2}{a4}
     \wordgroup[group style={fill=orange!70, draw=black, inner sep=.2ex}]{11}{3}{3}{a5}
     \wordgroup[group style={fill=orange!70, draw=black, inner sep=.2ex}]{11}{4}{4}{a6}
     \wordgroup[group style={fill=orange!70, draw=black, inner sep=.2ex}]{11}{5}{5}{a7}

\end{dependency}
\end{adjustbox}
\caption{Types of encoding on an example tree.}
\label{fig:dependecyTree}
\end{figure}

Thus, to encode a dependency tree, it suffices to encode the unique head position and dependency label associated with each word of $w_1 \ldots w_n$. To do so, we will give each word $w_i$ a discrete label of the form $(x_i,l_i)$, where $l_i$ is the dependency label and $x_i$ encodes the position of the head in one of the following four ways (see also Figure \ref{fig:dependecyTree}):

\begin{enumerate}[leftmargin=4mm,itemsep=1mm]
    \item Naive positional encoding: $x_i$ directly stores the 
    position of the head, 
    i.e., a label $(x_i,l_i)$ encodes an edge $(x_i,i,l_i)$. This is the encoding %commonly 
    used in the CoNLL file format.
    \item Relative positional encoding: $x_i$ stores the difference between the head index minus that of the dependent, i.e., $(x_i,l_i)$ encodes an edge $(i+x_i,i,l_i)$. This was the encoding used for the sequence-to-sequence and sequence labeling models in \citep{li-EtAl:2018:C18-13}, as well as for the sequence-to-sequence model in \citep{KiperwasserBallesteros18}.
    \item Relative PoS-based encoding: $x_i$ is a tuple $p_i,o_i$. If $o_i>0$,
    the head of $w_i$ is the $o_i$th closest among the words to the right of $w_i$ that have PoS tag $p_i$. 
    If $o_i < 0$,
    the head of $w_i$ is the $-o_i$th closest among the words to the left of $w_i$ that have PoS tag $p_i$. For example, $(V,-2)$ means ``the second verb to the left'' of $w_i$. 
    This scheme is closer to the notion of valency, and was used by \citet{Spoustova}.
    \item Bracketing-based encoding: based on 
    \cite{Yli2012,YliGomACL2017}.
    In each label $(x_i,l_i)$, the component $x_i$ is a string 
    following the regular expression
    \verb@(<)?((\)*|(/)*)(>)?@
    where the presence of character \verb@<@ means that $w_{i-1}$ has an incoming arc from the right, $k$ copies of character \verb@\@ mean that $w_{i}$ has $k$ outgoing arcs towards the left, $k$ copies of \verb@/@ mean that $w_{i-1}$ has $k$ outgoing arcs towards the right, and the presence of \verb@>@ means that $w_{i}$ has an incoming arc from the left. Thus, each right dependency from a word $i$ to $j$ is encoded by a $(\verb@/@,\verb@>@)$ pair in the label components $x_{i+1}$ and $x_j$, and each left dependency from $j$ to $i$ by a $(\verb@<@,\verb@\@)$ pair in the label components $x_{i+1}$ and $x_j$. Note that the intuition that explains why information related to a word is encoded in a neighboring node is that each $x_i$ corresponds to a fencepost position (i.e., $x_i$ represents the space between $w_{i-1}$ and $w_i$), and the character pair associated to an arc is encoded in the most external fencepost positions covered by that arc. These pairs act as pairs of matching brackets, which can be decoded using a stack to reconstruct the dependencies. 
\end{enumerate}

The first three encodings can represent any dependency tree, as they encode any valid head position for each node, while the bracketing encoding only supports projective trees, as it assumes that brackets are properly nested. 
All the encodings are total and injective,
but they are not surjective:
 head indexes can be out of range in the first three encodings, brackets can be unbalanced in encoding 4, and all the encodings can generate graphs with cycles. We will deal with ill-formed trees later.

\section{Model}

We use a standard encoder-decoder network, to show that dependency parsing as sequence labeling works without the need of complex models.

\paragraph{Encoder} We use bidirectional LSTMs \cite{hochreiter1997long,schuster1997bidirectional}.
Let LSTM$_\theta(\vec{x})$ be an abstraction of a long short-term memory network that processes the sequence of vectors $\vec{x}=[\vec{x}_1,...,\vec{x}_{|\vec{x}|}]$, then output for $\vec{x}_i$ is defined as $\vec{h}_i$ = BiLSTM$_\theta(\vec{x},i)$ = LSTM$_\theta^l(\vec{x}_{[1:i]}) \circ$ LSTM$_\theta^r(\vec{x}_{[|\vec{x}|:i]})$. We consider stacked BiLSTMs, where the output $\vec{h}^m_i$ of the $m$th BiLSTM layer is fed as input to the $m$+1th layer. Unless otherwise specified, the input token at a given time step is the concatenation of a word, PoS tag, and another word embedding learned through a character LSTM.

\paragraph{Decoder} 
We use a feed-forward network, which is fed the output of the last BiLSTM. The output is computed as $P(y_i|\vec{h}_i)$ = $\text{softmax}(W \cdot \vec{h}_i + \vec{b})$.

\paragraph{Well-formedness} (i) Each token must be assigned a head (one must be the dummy root), and (ii) the graph must be acyclic. If no token is the real root (no head is the dummy root), we search for candidates by relying on the three most likely labels for each token.\footnote{If single-rooted trees are a prerequisite, the most probable node will be selected among multiple root nodes.} If none is found, we assign it to the first token of the sentence. The single-head constraint is ensured by the nature of 
the encodings themselves,
but some of the predicted head indexes might be out of bounds. If so, we attach those tokens to the real root. If a cycle exists, we 
do the same for the leftmost token in the cycle.

\section{Experiments}

We use the English Penn Treebank (PTB) \cite{marcus1993building} and its splits for parsing.
We transform it into Stanford Dependencies \citep{deMarneffe} and obtain the predicted PoS tags using Stanford tagger \cite{tout}.
We also select a sample of UDv2.2 treebanks \citep{ud}: Ancient-Greek\textsubscript{PROIEL}, Czech\textsubscript{PDT}, Chinese\textsubscript{GSD}, English\textsubscript{EWT}, Finnish\textsubscript{TDT}, Hebrew\textsubscript{HTB}, Kazakh\textsubscript{KTB} and Tamil\textsubscript{TTB}, as a representative sample, following \citep{choiceOfUd}.
\noindent As evaluation metrics, we use Labeled (LAS) and Unlabeled Attachment Score (UAS).
We measure speed in sentences/second, both on a single core of a CPU\footnote{Intel Core i7-7700 CPU 4.2 GHz.} and on a GPU\footnote{GeForce GTX 1080.}.

\paragraph{Setup} 
We use NCRFpp as our sequence labeling framework \cite{yang2017ncrf}. For PTB, we use the embeddings by \newcite{emb}, for comparison to BIST parser \cite{TACL885}, which uses a similar architecture, but also needs a parsing algorithm and auxiliary structures. For UD, we follow an end-to-end setup and run UDPipe\footnote{The pretrained models from the CoNLL18 Shared Task.} \citep{udpipe:2017} for tokenization and tagging. We use the pretrained word embeddings by \newcite{embed}. 
Appendix \ref{appendix-training-configuration} contains additional hyperparameters.

\subsection{Encoding evaluation and model selection}

We first examine the four encodings on the PTB dev set.
Table \ref{tab:type-of-encoding} shows the results and also compares them against \citet{li-EtAl:2018:C18-13}, who proposed seq2seq and sequence labeling models that use a relative positional encoding.

\begin{table}[tbp]
\centering
\small{
\begin{tabular}{@{}lll@{}}
\toprule
Encoding & UAS & LAS\\
\midrule
\newcite{li-EtAl:2018:C18-13} (sequence labeling) & 87.58 & 83.81\\
\newcite{li-EtAl:2018:C18-13} (seq2seq) & 89.16 & 84.99\\
\newcite{li-EtAl:2018:C18-13} (seq2seq+beam+subroot) & 93.84 &\bf 91.86\\
\midrule
Naive positional& 45.41 & 42.65 \\
Rel. positional & 91.05 & 88.67 \\
Rel. PoS-based & \bf 93.99 &91.76 \\
Bracketing-based &  93.45 &  91.17 \\ 
\bottomrule
\end{tabular}}
\caption{Performance of our encodings on the PTB dev set with hyperparameters from \citet{GomVilEMNLP2018}. We compare against previous sequence labeling and seq2seq models with more complex architectures, beam search and subroot decomposition.}
\label{tab:type-of-encoding}
\end{table}

As the relative PoS-based encoding and bracketing-based encoding provide the best results, we will conduct the rest of our experiments with these two encodings.
Furthermore, we perform a small hyperparameter search involving encoding, number of hidden layers, their dimension and presence of character embeddings, as these parameters influence speed and accuracy.
From now on, we write $P^{\mathrm{z}}_{\mathrm{x,y}}$ for a PoS-based encoding model and $B^{\mathrm{z}}_{\mathrm{x,y}}$ for a bracketing-based encoding model, where $z$ indicates whether character representation was used in the model, $x$ the number of BiLSTM layers, and $y$ the word hidden vector dimension.
%For this purpose, 
We take as starting points (1) the hyperparameters used by the BIST parser \citep{TACL885}, as it uses a BiLSTM architecture analogous to ours, with the difference that it employs a transition-based algorithm that uses a stack data structure instead of plain sequence labeling without explicit representation of structure, and (2) the best hyperparameters used by \citet{GomVilEMNLP2018} for constituent parsing as sequence labeling, as it is an analogous task for a different parsing formalism. 

\begin{figure}[tbp]
    \centering
    \includegraphics[max width=\columnwidth]{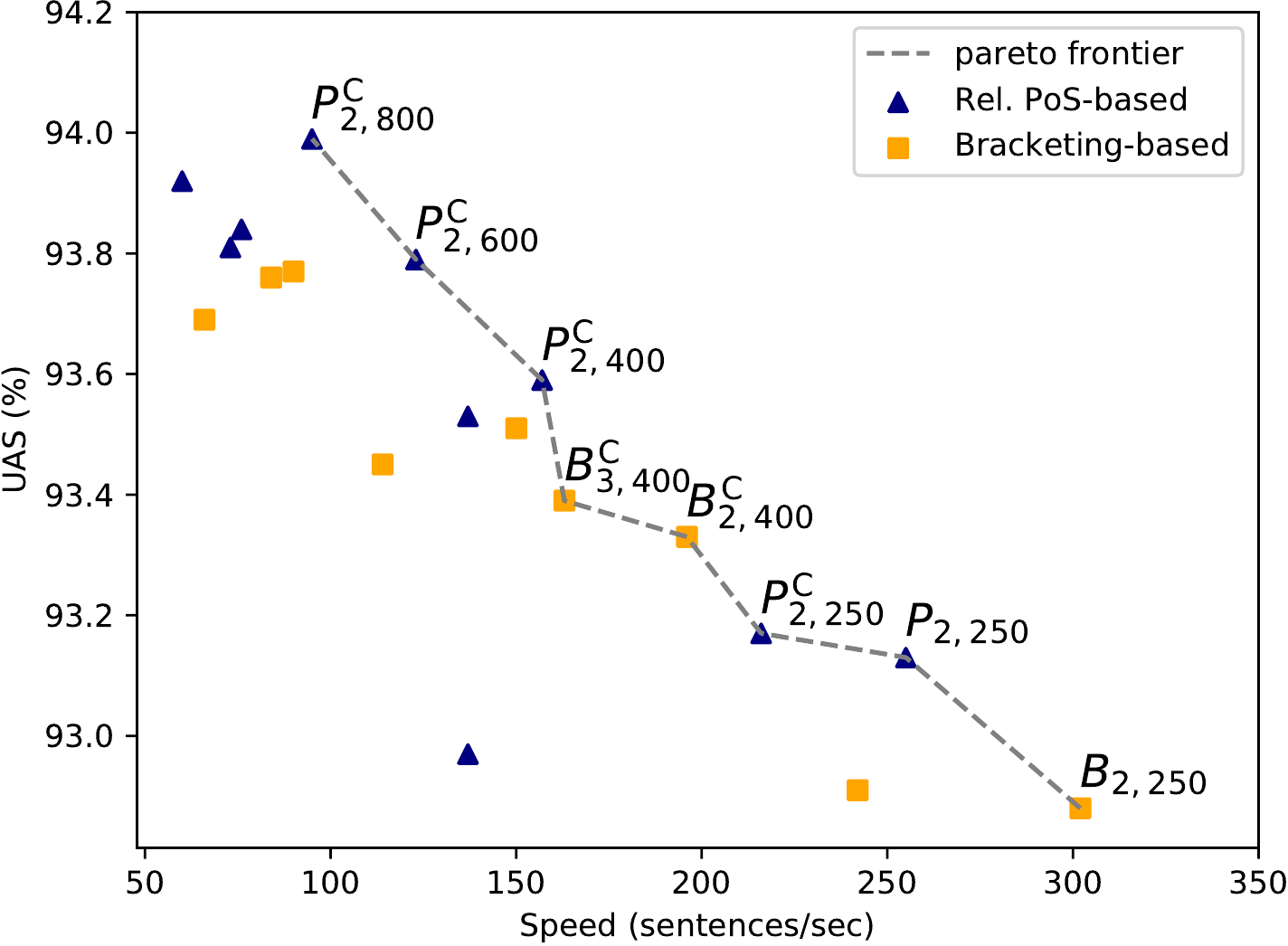}
    \caption{UAS/speed Pareto front on the PTB dev set.
    }
    \label{fig:pareto-uas}
\end{figure}

From there, we explore different combinations of parameters and evaluate 20 models on the PTB development set, with respect to accuracy (UAS) and speed (sentences/second on a single CPU core), obtaining the Pareto front in Figure \ref{fig:pareto-uas}. The two starting models based on previous literature ($P_{2,250}$ and $P^C_{2,800}$, respectively) happen to be in the Pareto front, confirming that they are reasonable hyperparameter choices also for this setting.
{In addition, we select two more models from the Pareto front (models $P^{\mathrm{C}}_{\mathrm{2,400}}$ and $B_{\mathrm{2,250}}$) for our test set experiments on PTB, as they also provide a good balance between speed and accuracy.}

\begin{table}[tbp]
\centering
\tabcolsep=0.15cm
\small{
\begin{tabular}{lcccc}
\hline
\multirow{2}{*}{Model}&\multicolumn{2}{c}{sent/s}& \multirow{2}{*}{ UAS}& \multirow{2}{*}{LAS} \\
& CPU &  GPU & &\\ 
\hline
$P_{\mathrm{2,250}}$  & \stddev{267}{1} & \stddev{777}{24} &  92.95 & 90.96\\
$P^{\mathrm{C}}_{\mathrm{2,400}}$  & \stddev{165}{1} & \stddev{700}{5} & 93.34 & 91.34 \\
$P^{\mathrm{C}}_{\mathrm{2,800}}$  & \stddev{101}{2} &  \stddev{648}{20} & \bf 93.67 & \bf 91.72\\
$B_{\mathrm{2,250}}$ & \stddev{310}{30} & \stddev{730}{53} & 92.64 & 90.59\\
\midrule
KG (transition-based) & \stddev{76}{1} & &    93.90 & 91.90 \\
KG (graph-based) & \stddev{80}{0} & &   93.10 & 91.00 \\
CM &  654$^\diamond$&&  91.80 & 89.60\\
DM&  &411$^\diamond$  & 95.74 & 94.08 \\ 
\newcite{MaStackPointer}&&\stddev{10}{0}&95.87&94.19\\
\hline
\end{tabular}
}
\caption{Comparison of models on the PTB test set. KG refers to \newcite{TACL885}, CM to \newcite{chen-manning:2014:EMNLP2014} and DM to \newcite{dozat-iclr}. $\diamond$ indicates the speed is taken from their paper. 
}
\label{tab:main_results}
\end{table}

\begin{table}[]
\begin{adjustbox}{max width=\columnwidth}
\begin{tabular}{@{}lccc|ccc@{}}
\hline
\multicolumn{1}{c}{} & \multicolumn{3}{c}{UPoS-based} & \multicolumn{3}{c}{XPoS-based} \\
Treebank & UAS & LAS & \# UPoS & UAS & LAS & \# XPoS \\ \hline
Ancient Greek & 76.58 & 71.70 & 14  & \textbf{77.00} & \textbf{72.14} & 23 \\
Chinese &  \textbf{61.01}& \textbf{57.28} & 15 & 60.98 & 57.14 & 42 \\
Czech & \textbf{89.82} & \textbf{87.63} & 17 & 88.33 & 85.46 &1417  \\
English & \textbf{82.22} & \textbf{78.96} & 17 & 82.05 & 78.70 & 50 \\
Finnish & \textbf{80.31} & \textbf{76.39} & 15 &80.19 & 76.28 & 12 \\
Hebrew\tablefootnote{In Hebrew, UPoS and XPoS tags are the same.} & 67.23 & 62.86 & 17 & 67.23 & 62.86 & 17 \\
Kazakh\tablefootnote{Kazakh is missing a development set. The scores are based on the test set.} & 32.14 & 17.03 & 15 & \textbf{32.93} & \textbf{17.07} & 26 \\
Tamil & \textbf{73.24} & \textbf{66.51} & 13 & 59.70 & 52.57 & 210 \\ \bottomrule
\end{tabular}
\end{adjustbox}
\caption{Performance of the $P^{\mathrm{C}}_{\mathrm{2,800}}$  model with UPoS- and XPoS-based encoding for each language on the dev set. 
\# UPoS/XPoS represents the number of distinct UPoS/XPoS tags in the training set for each language.
}
\label{tab:postags}
\end{table}

\begin{table}[tbp]
\tabcolsep=0.13cm
\begin{adjustbox}{max width=\columnwidth}
\small{
\begin{tabular}{l|l|ccc|ccc}
\hline
\multirow{3}{*}{\textbf{Treebank}} & &\multicolumn{3}{c|}{ $\mathbf{P^{C}_{2,800}}$}& \multicolumn{3}{c}{\textbf{KG (transition-based)}} \\
 & PoS type& \textbf{(sent/s)} & \multirow{2}{*}{ \textbf{UAS}} & \multirow{2}{*}{\textbf{LAS}}& \textbf{(sent/s)} & \multirow{2}{*}{\textbf{UAS}} &\multirow{2}{*}{ \textbf{LAS}}\\
 & &\textbf{CPU} &  &  & \textbf{CPU} & &  \\ \hline
 
Ancient Greek & XPOS&\stddev{123}{1} &   75.31  & 70.87 & \stddev{116}{4} & 69.43 & 64.41 \\
Chinese& UPOS &\stddev{105}{0}&  63.20 & 59.12 &\stddev{73}{1}&  64.69 & 60.45 \\
Czech& UPOS &\stddev{125}{1}&  89.10  & 86.68  & \stddev{94}{3}& 89.25 &86.11  \\
English& UPOS &\stddev{139}{1} &  81.48& 78.64 &\stddev{120}{2}&   82.22 & 79.00 \\
Finnish& UPOS &\stddev{168}{0}& 80.12 & 76.22   &\stddev{127}{3} &80.99 & 76.63     \\
Hebrew & equal PoS&\stddev{120}{0} &  63.04 & 58.66  & \stddev{70}{1}& 63.56 & 58.80   \\
Kazakh &XPOS&\stddev{283}{3} &32.93&  17.07  & \stddev{178}{5} & 23.09& 12.73   \\
Tamil\tablefootnote{Tamil was run on gold segmented and tokenized inputs, as there is no pretrained UDpipe model. We did not use pretrained word embeddings either.} & UPOS& \stddev{150}{2} & 71.59 & 64.00 &\stddev{127}{3} & 75.41 & 68.58   \\
\hline
\end{tabular}}
\end{adjustbox}
\caption{Comparison on UD-CoNLL18 test sets.}
\label{tab:ud_results}
\end{table}

\begin{figure}[tbp]
    \centering
    \includegraphics[max width=\columnwidth]{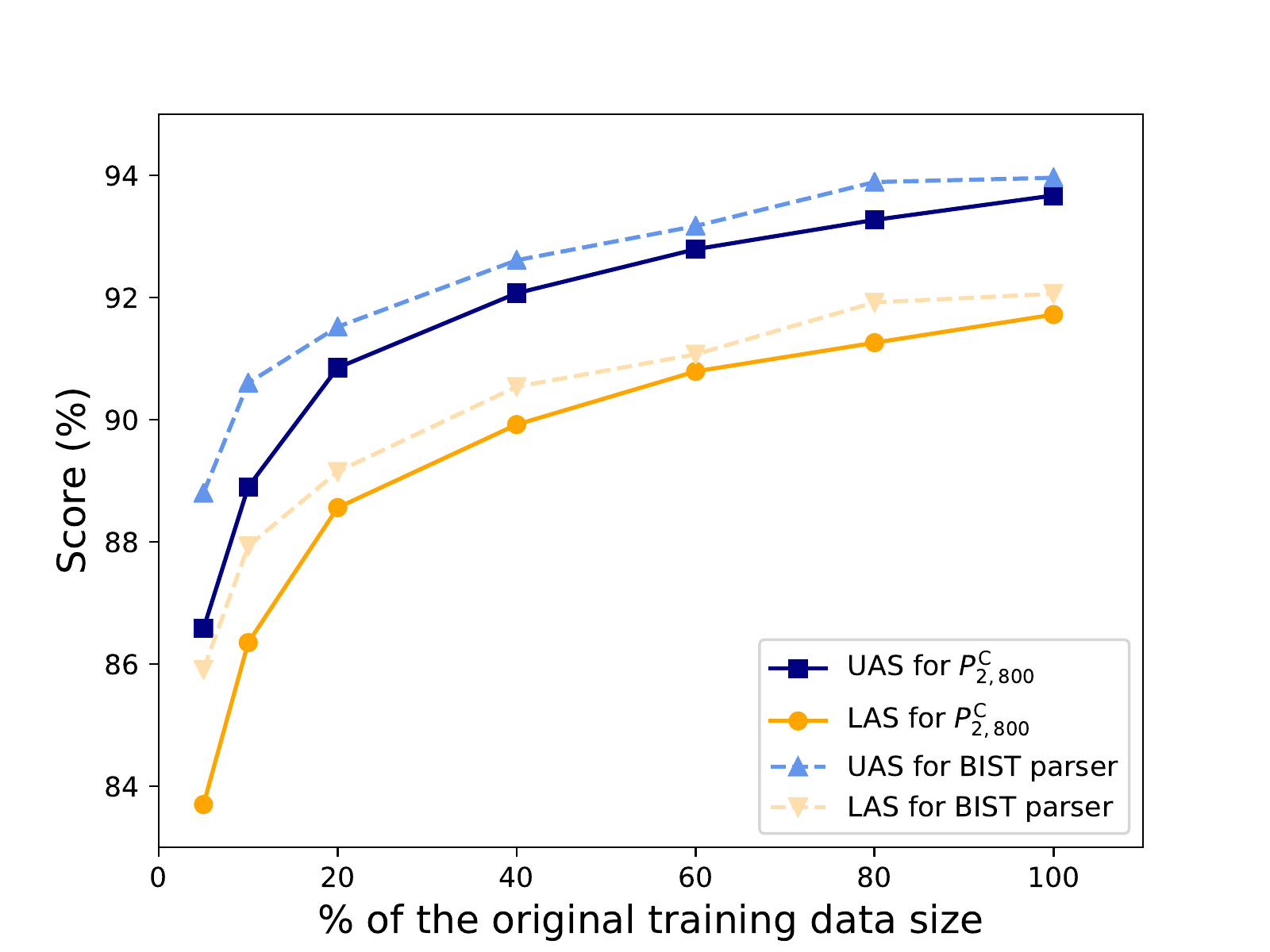}
    \caption{Impact of the PTB data size available for parsers during training on the results from the test set. }
    \label{fig:dataSize}
\end{figure}

\subsection{Results and discussion}

Table \ref{tab:main_results} compares the chosen models, on the PTB test set, against state-of-the-art models. Contrary to previous dependency-parsing-as-sequence-labeling attempts, we are competitive and provide a good speed-accuracy tradeoff. For instance, the $P^{\mathrm{C}}_{\mathrm{2,800}}$ model runs faster than 
the BIST parser \citep{TACL885} while being almost as accurate (-0.18 LAS). This comes in spite of its simplicity. While our BiLSTM architecture is similar to that of BIST,
the sequence labeling approach does not need a stack, a specific transition system or a dynamic oracle. Using the BIST hyperparameters for our model ($P_{\mathrm{2,250}}$) yields further increases in speed, at some cost to accuracy: 3.34x faster and -0.04 LAS score than the graph-based model, and 3.51x faster and -0.94 LAS score than their transition-based one.

We now extend our experiments to the sample of UD-CoNLL18 treebanks. To this end, we focus on the $P^{C}_{2,800}$ model and since our PoS tag-based encoding can be influenced by the specific PoS tags used, we first conduct an experiment on the development sets to determine what tag set (UPoS, the universal PoS tag set, common to all languages, or XPoS, extended language-specific PoS tags) produces the best results for each dataset. 

Table \ref{tab:postags} shows how the number of unique UPoS and XPoS tags found in the training set differs in various languages. The results suggest that the performance of our system can be influenced by the size of the tag set. It appears that a very large tag set (for instance the XPoS tag set for Czech and Tamil) can hurt the performance of the model and significantly slow down the system, as it results into a large number of distinct labels for the sequence labeling model, increasing sparsity and making the classification harder. In case of Ancient Greek and Kazakh, the best performance is achieved with the XPoS-based encoding. In these corpora, the tag set is slightly bigger than the UPoS tag set. One can argue that the XPoS tags in this case were possibly more fine-grained and hence provided additional useful information to the system facilitating a correct label prediction, without being so large as to produce excessive sparsity. 

Table \ref{tab:ud_results} shows experiments on the UD test sets, with the chosen PoS tag set for each corpus. $P^{\mathrm{C}}_{\mathrm{2,800}}$ outperforms transition-based BIST in LAS in 3 out of 8 treebanks,\footnote{For Ancient Greek, this may be related to the large amount of non-projectivity (BIST is a projective parser). For extra comparison, a non-projective variant of BIST \citep{UppsalaConll18} obtains 71.58 LAS with mono-treebank training, but from better segmentation and morphology than used here. UDpipe \citep{udpipe:2017} obtains 67.57 LAS.
Czech and Kazakh have a medium amount of non-projectivity.
}
and is clearly faster in all analyzed languages.
We believe that the variations between languages in terms of LAS difference with respect to BIST can be largely due to differences in the accuracy and granularity of predicted PoS tags,
since our chosen encoding relies on them to encode arcs.
The bracketing-based encoding, which does not use PoS tags, may be more robust to this. On the other hand, finding the optimal granularity of PoS tags for the PoS-based encoding can be an interesting avenue for future work.

In this work, we have also examined the impact of the training data size on the performance of our system compared to the performance of BIST parser. The results in Figure \ref{fig:dataSize} suggest that our model requires more data during the training than BIST parser in order to achieve similar performance. The performance is slightly worse when little training data is available, but later on our model reduces the gap when increasing the training data size.

\section{Conclusion}

This paper has explored fast and accurate dependency parsing as sequence labeling. We tested four different encodings, training a standard BiLSTM-based architecture. In contrast to previous work, our results on the PTB and a subset of UD treebanks show that this paradigm can obtain competitive results, despite not using any parsing algorithm nor external structures to parse sentences.

\section*{Acknowledgments}

This work has received funding from the European
Research Council (ERC), under the European
Union's Horizon 2020 research and innovation
programme (FASTPARSE, grant agreement No
714150), from the TELEPARES-UDC project
(FFI2014-51978-C2-2-R) and the ANSWER-ASAP project (TIN2017-85160-C2-1-R) from MINECO, and from Xunta de Galicia (ED431B 2017/01). We gratefully acknowledge NVIDIA Corporation for the donation of a GTX Titan X GPU.

\bibliographystyle{acl_natbib}
\bibliography{naaclhlt2019}

\begin{thebibliography}{34}
\expandafter\ifx\csname natexlab\endcsname\relax\def\natexlab#1{#1}\fi

\bibitem[{Brill(1995)}]{brill1995transformation}
Eric Brill. 1995.
\newblock Transformation-based error-driven learning and natural language
  processing: A case study in part-of-speech tagging.
\newblock \emph{Computational linguistics}, 21(4):543--565.

\bibitem[{Chen and Manning(2014)}]{chen-manning:2014:EMNLP2014}
Danqi Chen and Christopher~D. Manning. 2014.
\newblock \href {http://www.aclweb.org/anthology/D14-1082} {A fast and accurate
  dependency parser using neural networks}.
\newblock In \emph{Proceedings of the 2014 Conference on Empirical Methods in
  Natural Language Processing (EMNLP)}, pages 740--750, Doha, Qatar.
  Association for Computational Linguistics.

\bibitem[{De~Marneffe et~al.(2006)De~Marneffe, MacCartney, Manning
  et~al.}]{deMarneffe}
Marie-Catherine De~Marneffe, Bill MacCartney, Christopher~D Manning, et~al.
  2006.
\newblock Generating typed dependency parses from phrase structure parses.
\newblock In \emph{Lrec}, volume~6, pages 449--454.

\bibitem[{Dozat and Manning(2017)}]{dozat-iclr}
Timothy Dozat and Christopher~D. Manning. 2017.
\newblock Deep biaffine attention for neural dependency parsing.
\newblock In \emph{Proceedings of the 5th International Conference on Learning
  Representations}.

\bibitem[{Dyer et~al.(2015)Dyer, Ballesteros, Ling, Matthews, and
  Smith}]{Dyer2015}
Chris Dyer, Miguel Ballesteros, Wang Ling, Austin Matthews, and Noah~A. Smith.
  2015.
\newblock \href {https://doi.org/10.3115/v1/P15-1033} {Transition-based
  dependency parsing with stack long short-term memory}.
\newblock In \emph{Proceedings of the 53rd Annual Meeting of the Association
  for Computational Linguistics and the 7th International Joint Conference on
  Natural Language Processing (Volume 1: Long Papers)}, pages 334--343.
  Association for Computational Linguistics.

\bibitem[{Ginter et~al.(2017)Ginter, Haji{\v c}, Luotolahti, Straka, and
  Zeman}]{embed}
Filip Ginter, Jan Haji{\v c}, Juhani Luotolahti, Milan Straka, and Daniel
  Zeman. 2017.
\newblock \href {http://hdl.handle.net/11234/1-1989} {{CoNLL} 2017 shared task
  - automatically annotated raw texts and word embeddings}.
\newblock {LINDAT}/{CLARIN} digital library at the Institute of Formal and
  Applied Linguistics ({{\'U}FAL}), Faculty of Mathematics and Physics, Charles
  University.

\bibitem[{G{\'o}mez-Rodr{\'i}guez and Vilares(2018)}]{GomVilEMNLP2018}
Carlos G{\'o}mez-Rodr{\'i}guez and David Vilares. 2018.
\newblock \href {http://aclweb.org/anthology/D18-1162} {Constituent parsing as
  sequence labeling}.
\newblock In \emph{Proceedings of the 2018 Conference on Empirical Methods in
  Natural Language Processing}, pages 1314--1324. Association for Computational
  Linguistics.

\bibitem[{Hochreiter and Schmidhuber(1997)}]{hochreiter1997long}
Sepp Hochreiter and J{\"u}rgen Schmidhuber. 1997.
\newblock Long short-term memory.
\newblock \emph{Neural computation}, 9(8):1735--1780.

\bibitem[{Huang et~al.(2015)Huang, Xu, and Yu}]{huang2015bidirectional}
Zhiheng Huang, Wei Xu, and Kai Yu. 2015.
\newblock Bidirectional {LSTM-CRF} models for sequence tagging.
\newblock \emph{arXiv preprint arXiv:1508.01991}.

\bibitem[{Kiperwasser and Ballesteros(2018)}]{KiperwasserBallesteros18}
Eliyahu Kiperwasser and Miguel Ballesteros. 2018.
\newblock \href {http://www.aclweb.org/anthology/Q18-1017} {Scheduled
  multi-task learning: From syntax to translation}.
\newblock \emph{Transactions of the Association for Computational Linguistics},
  6:225--240.

\bibitem[{Kiperwasser and Goldberg(2016)}]{TACL885}
Eliyahu Kiperwasser and Yoav Goldberg. 2016.
\newblock \href {https://transacl.org/ojs/index.php/tacl/article/view/885}
  {Simple and accurate dependency parsing using bidirectional {LSTM} feature
  representations}.
\newblock \emph{Transactions of the Association for Computational Linguistics},
  4:313--327.

\bibitem[{Koo and Collins(2010)}]{koo-collins:2010:ACL}
Terry Koo and Michael Collins. 2010.
\newblock \href {http://www.aclweb.org/anthology/P10-1001} {Efficient
  third-order dependency parsers}.
\newblock In \emph{Proceedings of the 48th Annual Meeting of the Association
  for Computational Linguistics}, pages 1--11, Uppsala, Sweden. Association for
  Computational Linguistics.

\bibitem[{de~Lhoneux et~al.(2017)de~Lhoneux, Stymne, and Nivre}]{choiceOfUd}
Miryam de~Lhoneux, Sara Stymne, and Joakim Nivre. 2017.
\newblock Old school vs. new school: Comparing transition-based parsers with
  and without neural network enhancement.
\newblock In \emph{TLT}, pages 99--110.

\bibitem[{Li et~al.(2018)Li, Cai, He, and Zhao}]{li-EtAl:2018:C18-13}
Zuchao Li, Jiaxun Cai, Shexia He, and Hai Zhao. 2018.
\newblock \href {http://www.aclweb.org/anthology/C18-1271} {Seq2seq dependency
  parsing}.
\newblock In \emph{Proceedings of the 27th International Conference on
  Computational Linguistics}, pages 3203--3214, Santa Fe, New Mexico, USA.
  Association for Computational Linguistics.

\bibitem[{Ling et~al.(2015)Ling, Dyer, Black, and Trancoso}]{emb}
Wang Ling, Chris Dyer, Alan~W Black, and Isabel Trancoso. 2015.
\newblock \href {https://doi.org/10.3115/v1/N15-1142} {Two/too simple
  adaptations of word2vec for syntax problems}.
\newblock In \emph{Proceedings of the 2015 Conference of the North American
  Chapter of the Association for Computational Linguistics: Human Language
  Technologies}, pages 1299--1304. Association for Computational Linguistics.

\bibitem[{Ma et~al.(2018)Ma, Hu, Liu, Peng, Neubig, and Hovy}]{MaStackPointer}
Xuezhe Ma, Zecong Hu, Jingzhou Liu, Nanyun Peng, Graham Neubig, and Eduard
  Hovy. 2018.
\newblock \href {http://aclweb.org/anthology/P18-1130} {Stack-pointer networks
  for dependency parsing}.
\newblock In \emph{Proceedings of the 56th Annual Meeting of the Association
  for Computational Linguistics (Volume 1: Long Papers)}, pages 1403--1414.
  Association for Computational Linguistics.

\bibitem[{Marcus et~al.(1993)Marcus, Marcinkiewicz, and
  Santorini}]{marcus1993building}
Mitchell~P Marcus, Mary~Ann Marcinkiewicz, and Beatrice Santorini. 1993.
\newblock Building a large annotated corpus of {E}nglish: The {P}enn treebank.
\newblock \emph{Computational linguistics}, 19(2):313--330.

\bibitem[{Nivre et~al.(2018)}]{ud}
Joakim Nivre et~al. 2018.
\newblock \href {http://hdl.handle.net/11234/1-2837} {Universal dependencies
  2.2}.
\newblock {LINDAT}/{CLARIN} digital library at the Institute of Formal and
  Applied Linguistics ({{\'U}FAL}), Faculty of Mathematics and Physics, Charles
  University.

\bibitem[{Ramshaw and Marcus(1999)}]{ramshaw1999text}
Lance~A Ramshaw and Mitchell~P Marcus. 1999.
\newblock Text chunking using transformation-based learning.
\newblock In \emph{Natural language processing using very large corpora}, pages
  157--176. Springer.

\bibitem[{Reimers and Gurevych(2017)}]{reimers2017reporting}
Nils Reimers and Iryna Gurevych. 2017.
\newblock \href {https://doi.org/10.18653/v1/D17-1035} {Reporting score
  distributions makes a difference: Performance study of lstm-networks for
  sequence tagging}.
\newblock In \emph{Proceedings of the 2017 Conference on Empirical Methods in
  Natural Language Processing}, pages 338--348. Association for Computational
  Linguistics.

\bibitem[{Schuster and Paliwal(1997)}]{schuster1997bidirectional}
Mike Schuster and Kuldip~K Paliwal. 1997.
\newblock Bidirectional recurrent neural networks.
\newblock \emph{IEEE Transactions on Signal Processing}, 45(11):2673--2681.

\bibitem[{Shi et~al.(2017)Shi, Huang, and Lee}]{shi-huang-lee:2017:EMNLP2017}
Tianze Shi, Liang Huang, and Lillian Lee. 2017.
\newblock \href {https://www.aclweb.org/anthology/D17-1002} {Fast(er) exact
  decoding and global training for transition-based dependency parsing via a
  minimal feature set}.
\newblock In \emph{Proceedings of the 2017 Conference on Empirical Methods in
  Natural Language Processing}, pages 12--23, Copenhagen, Denmark. Association
  for Computational Linguistics.

\bibitem[{Smith et~al.(2018)Smith, Bohnet, de~Lhoneux, Nivre, Shao, and
  Stymne}]{UppsalaConll18}
Aaron Smith, Bernd Bohnet, Miryam de~Lhoneux, Joakim Nivre, Yan Shao, and Sara
  Stymne. 2018.
\newblock \href {http://aclweb.org/anthology/K18-2011} {82 treebanks, 34
  models: Universal dependency parsing with multi-treebank models}.
\newblock In \emph{Proceedings of the CoNLL 2018 Shared Task: Multilingual
  Parsing from Raw Text to Universal Dependencies}, pages 113--123. Association
  for Computational Linguistics.

\bibitem[{Spoustov\'a and Spousta(2010)}]{Spoustova}
Drahom\'ira Spoustov\'a and Miroslav Spousta. 2010.
\newblock \href
  {https://content.sciendo.com/view/journals/pralin/94/1/article-p7.xml}
  {Dependency parsing as a sequence labeling task}.
\newblock \emph{The Prague Bulletin of Mathematical Linguistics}, 94(1):7--14.

\bibitem[{Straka and Strakov\'{a}(2017)}]{udpipe:2017}
Milan Straka and Jana Strakov\'{a}. 2017.
\newblock \href {http://www.aclweb.org/anthology/K/K17/K17-3009.pdf}
  {Tokenizing, pos tagging, lemmatizing and parsing ud 2.0 with udpipe}.
\newblock In \emph{Proceedings of the CoNLL 2017 Shared Task: Multilingual
  Parsing from Raw Text to Universal Dependencies}, pages 88--99, Vancouver,
  Canada. Association for Computational Linguistics.

\bibitem[{Toutanova et~al.(2003)Toutanova, Klein, Manning, and Singer}]{tout}
Kristina Toutanova, Dan Klein, Christopher~D Manning, and Yoram Singer. 2003.
\newblock Feature-rich part-of-speech tagging with a cyclic dependency network.
\newblock In \emph{Proceedings of the 2003 Conference of the North American
  Chapter of the Association for Computational Linguistics on Human Language
  Technology-Volume 1}, pages 173--180. Association for Computational
  Linguistics.

\bibitem[{Vinyals et~al.(2015)Vinyals, Kaiser, Koo, Petrov, Sutskever, and
  Hinton}]{vinyals2015grammar}
Oriol Vinyals, {\L}ukasz Kaiser, Terry Koo, Slav Petrov, Ilya Sutskever, and
  Geoffrey Hinton. 2015.
\newblock Grammar as a foreign language.
\newblock In \emph{Advances in Neural Information Processing Systems}, pages
  2773--2781.

\bibitem[{Wiseman and Rush(2016)}]{wiseman-rush:2016:EMNLP2016}
Sam Wiseman and Alexander~M. Rush. 2016.
\newblock \href {https://aclweb.org/anthology/D16-1137} {Sequence-to-sequence
  learning as beam-search optimization}.
\newblock In \emph{Proceedings of the 2016 Conference on Empirical Methods in
  Natural Language Processing}, pages 1296--1306, Austin, Texas. Association
  for Computational Linguistics.

\bibitem[{Yang and Zhang(2018)}]{yang2017ncrf}
Jie Yang and Yue Zhang. 2018.
\newblock \href {http://www.aclweb.org/anthology/P18-4013} {{NCRF}++: An
  open-source neural sequence labeling toolkit}.
\newblock In \emph{Proceedings of ACL 2018, System Demonstrations}, pages
  74--79, Melbourne, Australia. Association for Computational Linguistics.

\bibitem[{Yli-Jyr{\"a}(2012)}]{Yli2012}
Anssi Yli-Jyr{\"a}. 2012.
\newblock \href {https://doi.org/10.1007/978-3-642-30773-7_10} {\emph{On
  Dependency Analysis via Contractions and Weighted FSTs}}, pages 133--158.
  Springer Berlin Heidelberg, Berlin, Heidelberg.

\bibitem[{Yli-Jyr\"{a} and G\'{o}mez-Rodr\'{i}guez(2017)}]{YliGomACL2017}
Anssi Yli-Jyr\"{a} and Carlos G\'{o}mez-Rodr\'{i}guez. 2017.
\newblock \href {http://aclweb.org/anthology/P17-1160} {Generic axiomatization
  of families of noncrossing graphs in dependency parsing}.
\newblock In \emph{Proceedings of the 55th Annual Meeting of the Association
  for Computational Linguistics (Volume 1: Long Papers)}, pages 1745--1755,
  Vancouver, Canada. Association for Computational Linguistics.

\bibitem[{Zhang et~al.(2017{\natexlab{a}})Zhang, Cheng, and
  Lapata}]{zhang2017dependency}
Xingxing Zhang, Jianpeng Cheng, and Mirella Lapata. 2017{\natexlab{a}}.
\newblock Dependency parsing as head selection.
\newblock In \emph{Proceedings of the 15th Conference of the European Chapter
  of the Association for Computational Linguistics: Volume 1, Long Papers},
  volume~1, pages 665--676.

\bibitem[{Zhang and Nivre(2011)}]{zhang-nivre:2011:ACL-HLT2011}
Yue Zhang and Joakim Nivre. 2011.
\newblock \href {http://www.aclweb.org/anthology/P11-2033} {Transition-based
  dependency parsing with rich non-local features}.
\newblock In \emph{Proceedings of the 49th Annual Meeting of the Association
  for Computational Linguistics: Human Language Technologies}, pages 188--193,
  Portland, Oregon, USA. Association for Computational Linguistics.

\bibitem[{Zhang et~al.(2017{\natexlab{b}})Zhang, Liu, Li, Zhou, and
  Chen}]{zhang-EtAl:2017:EMNLP20173}
Zhirui Zhang, Shujie Liu, Mu~Li, Ming Zhou, and Enhong Chen.
  2017{\natexlab{b}}.
\newblock \href {https://www.aclweb.org/anthology/D17-1175} {Stack-based
  multi-layer attention for transition-based dependency parsing}.
\newblock In \emph{Proceedings of the 2017 Conference on Empirical Methods in
  Natural Language Processing}, pages 1677--1682, Copenhagen, Denmark.
  Association for Computational Linguistics.

\end{thebibliography}
\clearpage
\appendix

\section{Model parameters}\label{appendix-training-configuration}

During the training we use Stochastic Gradient Descent (SGD) optimizer with a batch size of 8, and the model is trained for up to 100 iterations. We keep the model that obtains the highest UAS on the development set. Additional hyperparameters are shown in Table~\ref{tab:hyper}.

\begin{table}[hbtp]
\begin{small}
\begin{tabular}{|l|l|}
\hline
Word embedding dimension & 100 \\
Char embedding dimension & 30 \\
PoS tag embedding dimension & 25 \\
Word hidden vector dimension & \begin{tabular}[c]{@{}l@{}}250,400,600,\\ 800,1000,1200\end{tabular} \\
Character hidden vector dimension & 50 \\
Initial learning rate & 0.02 \\
Time-based learning rate decay & 0.05 \\
Momentum & 0.9 \\
Dropout & 0.5 \\ \hline
\end{tabular}
\end{small}
\caption{Common hyperparameters for the sequence labeling models.}
\label{tab:hyper}
\end{table}

\end{document}